\documentclass[10pt,twocolumn,letterpaper]{article}

\usepackage{iccv}
\usepackage{times}
\usepackage{epsfig}
\usepackage{graphicx}
\usepackage{amsmath}
\usepackage{amssymb}

\usepackage{subfigure}
\usepackage{multirow}
\usepackage[cmyk]{xcolor}
\usepackage{booktabs}

\usepackage[breaklinks=true,bookmarks=false]{hyperref}

\iccvfinalcopy 

\def\iccvPaperID{4261} 

\ificcvfinal\fi

\begin{document}

\title{Multi-Label Self-Supervised Learning with Scene Images}

\author{Ke Zhu \quad Minghao Fu \quad Jianxin Wu\thanks{J. Wu is the corresponding author. This paper was partly supported by the National Natural Science Foundation of China under Grant 62276123 and Grant 61921006.}\\
State Key Laboratory for Novel Software Technology\\
Nanjing University, China \\
{\tt\small zhuk@lamda.nju.edu.cn, fumh@lamda.nju.edu.cn, wujx2001@nju.edu.cn}
}

\maketitle
\ificcvfinal\thispagestyle{empty}\fi

\begin{abstract}
   Self-supervised learning (SSL) methods targeting scene images have seen a rapid growth recently, and they mostly rely on either a dedicated dense matching mechanism or a costly unsupervised object discovery module. This paper shows that instead of hinging on these strenuous operations, quality image representations can be learned by treating scene/multi-label image SSL simply as a multi-label classification problem, which greatly simplifies the learning framework. Specifically, multiple binary pseudo-labels are assigned for each input image by comparing its embeddings with those in two dictionaries, and the network is optimized using the binary cross entropy loss. The proposed method is named Multi-Label Self-supervised learning (MLS). Visualizations qualitatively show that clearly the pseudo-labels by MLS can automatically find semantically similar pseudo-positive pairs across different images to facilitate contrastive learning. MLS learns high quality representations on MS-COCO and achieves state-of-the-art results on classification, detection and segmentation benchmarks. At the same time, MLS is much simpler than existing methods, making it easier to deploy and for further exploration.
\end{abstract}

\section{Introduction}\label{sec:intro}

Self-supervised learning (SSL) methods based on contrastive learning~\cite{SimCLR,SwAV} have facilitated numerous downstream tasks. Those~\cite{MOCO,MOCOv2,BYOL,SSL_in_the_wild} that target object-centric images (\eg, ImageNet) are already relatively mature, but inventing SSL methods for scene images (\eg, MS-COCO~\cite{MS-COCO}) gains popularity recently. Since unlabeled scene images (or multi-label images) are more natural~\cite{ORL} and richer in semantics, various SSL methods~\cite{DenseCL,DetCo,DetCon,SetSim,LEWEL,ReSim} have successively emerged.

SSL methods focusing on scene/multi-label images can be summarized into two categories. One is \emph{dense matching}, such as DenseCL~\cite{DenseCL}, MaskCo~\cite{MaskCo} and Self-EMD~\cite{self-EMD}. They take the features' locations into account to improve the performance on dense prediction tasks. They mostly differ in how the heuristic matching metric is designed. Another branch of work like SoCo~\cite{SoCo} and ORL~\cite{ORL} resort to \emph{unsupervised object discovery} to find local object contents, and learn quality representation with both object- and scene-level semantics. However, they usually involve multi-stage SSL pretraining on top of expensive box generation~\cite{SS}.

\begin{figure}
  \centering
  \includegraphics[width=0.9\linewidth]{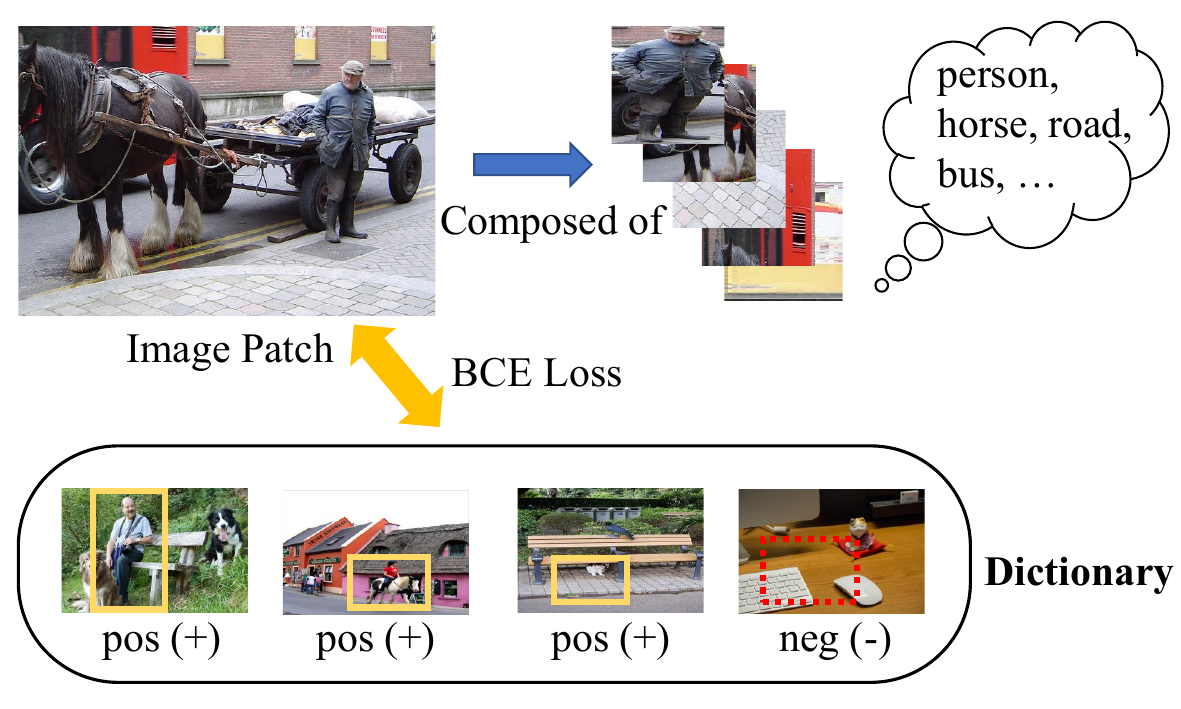}
  \caption{Illustrating our motivation. An image patch cropped from a multi-label image comprises multiple objects. Each object can find similar (`pos') and dissimilar (`neg') images from a large dictionary. The whole image patch is pulled closer to those positive ones and pushed away from the negatives using a BCE loss. See~Fig.~\ref{fig:crosspondence_Visualization} for positives and negatives chosen by our algorithm.}
  \label{fig:figure1}
\end{figure}

These scene image SSL methods are generally based on the contrastive loss (\eg, InfoNCE~\cite{InfoNCE}), where two randomly augmented views of the same image are forced to be close to each other, and \emph{optionally} push away views from different images. The loss assumes single-label images~\cite{InfoNCE}, but the input are in fact multi-label: hence \emph{there is a mismatch between the loss and the data}. On one hand, it can be difficult for two views \emph{randomly} cropped from the same scene image to be exactly matched~\cite{DenseCL,PixPro}. On the other hand, there is only \emph{one} matched positive pair in InfoNCE, while more positive pairs are naturally preferred: \emph{a view cropped from a multi-label image likely contains multiple semantic concepts or objects}.

Therefore, this paper proposes a simple yet direct approach towards scene image SSL, named as Multi-Label Self-supervised learning (MLS). We treat each image (or randomly cropped patch) as a semantic bag with multiple objects, then retrieve images sharing similar semantics with any object in the bag from a large image dictionary. Note that an object in a bag is not necessarily within the set of human-annotated categories. As illustrated in Fig.~\ref{fig:figure1}, the cropped patch contains \emph{person}, \emph{horse}, \emph{road} and \emph{bus}, which will be pulled closer to similar images containing any of these objects and be pushed away from those dissimilar ones. Specifically, the patch's embedding produced by a backbone network will select top $k$ similar embeddings as $k$ (pseudo) positive images from a dictionary, and the rest will be negatives. In another dictionary containing images in the same order as the first one, the BCE (binary cross entropy) loss plus these binary pseudo-labels will classify the patch’s embedding after an MLP projector using all the images in this second dictionary as classifiers, and generates gradients that optimize the backbone network. This framework is illustrated in Fig.~\ref{fig:figure2}.

Our framework has two benefits. First, the large dictionary has diverse positive samples for any given input, hence provides many quality positive pairs with deformations or intra-class variations~\cite{NNCLR}. Second, unlike InfoNCE, the BCE loss is \emph{not} mutually exclusive among classes, hence \emph{allows the co-occurrence of multiple classes in one scene image}. By applying our SSL method to scene images (\eg, MS-COCO), we achieve state-of-the-art results on object detection, instance segmentation and various classification benchmarks. Our contributions are summarized as follows:

\begin{enumerate}
 \item For the first time, we formulate scene image SSL as a multi-label classification, and propose our Multi-Label Self-supervised (MLS) learning approach.
 \item Unlike previous methods that adopt dense matching or unsupervised object discovery, MLS is simple in concept, and enjoys intuitive visualizations (\cf Fig.~\ref{fig:crosspondence_Visualization}) which clearly verifies our motivation.
 \item Extensive experiments of object detection, instance segmentation and classification on various benchmark datasets, together with ablation studies, clearly demonstrate the effectiveness of our method.
\end{enumerate}

\section{Related Work}

\textbf{Traditional SSL methods.} Self-supervised learning (SSL) has emerged as a promising direction towards unsupervised representation learning~\cite{MOCOv2}. Early SSL methods~\cite{InfoNCE,CMC} are mainly derived from noise contrastive estimation~\cite{NCE}. Later, SSL methods based on contrastive learning and clustering paradigms have been proved effective, too. Representative methods such as SimCLR~\cite{SimCLR}, MoCo~\cite{MOCO}, BYOL~\cite{BYOL} and SwAV~\cite{SwAV} have exhibited both simplicity and generalization ability~\cite{NIPs_revisiting} to various downstream tasks. A set of variants such as  Simsiam~\cite{Simsiam}, InfoMin~\cite{InfoMin} and NNCLR~\cite{NNCLR} try to simplify, analyze and optimize traditional SSL methods from different aspects. However, all of them mainly focus on pretraining on object-centric images like the ImageNet dataset~\cite{ImageNet} for image classification (\eg, ImageNet linear evaluation), but pay less attention to dense prediction tasks.

\textbf{Scene-image SSL methods.} The development of scene (multi-label) image SSL is a diversified process~\cite{DenseCL,Cropping}, which mainly contains two branches: dedicated dense feature matching and multi-stage pretraining with unsupervised object discovery. DenseCL~\cite{DenseCL}, PixPro~\cite{PixPro}, ReSim~\cite{ReSim}, LEWEL~\cite{LEWEL} and SetSim~\cite{SetSim} all apply the InfoNCE loss in a dense matching manner. They propose dense loss functions in addition to the traditional InfoNCE loss, and their pipelines differ in how the matching metrics are (probably manually) selected. Another branch of methods such as SoCo~\cite{SoCo} and ORL~\cite{ORL} try to learn quality feature representation with both scene- and object-level information: they utilize multi-stage pretraining to find correspondence across images with the help of unsupervised object generation methods~\cite{SS,EdgeBox}, which suffers from huge computation costs. A recent SSL method~\cite{NIPs_revisiting} that is close to our work proposed kNN-MoCo as an extra module. But, this method is still in a single-label InfoNCE manner and the improvement is small compared to our multi-label MLS (\cf Table~\ref{tab:different_bce}). Compared to existing methods, our MLS is novel in concept and simple in implementation.

\textbf{Multi-label classification.} Multi-label recognition aims to predict the presence or absence of each object class in an input image~\cite{multi-label-review}. Traditional methods are based on three aspects: attention modules~\cite{CSRA,SRN}, correlation matrix~\cite{GCN,GCNre} and unsupervised box generation~\cite{2015_PAMI_HCP,2016_ICIP_RCP}. The loss function adopted in this task is mostly binary cross entropy (BCE), which is \emph{not} mutually exclusive among classes~\cite{multi-label-review}, such that the presence of one class does not suppress the existence of others. In this paper, we inherit this unique property of multi-label classification and try to integrate BCE into self-supervised learning on scene images. We adopt the joint image embedding to generate $k$ pseudo-labels, together with the classification logits to calculate the final loss term, which effectively guides the multi-label self-supervised learning.

\section{Method}

In this section, we start by describing the preliminaries of self-supervised learning, including the InfoNCE loss and its variants. We will then describe the proposed Multi-Label Self-supervised (MLS) learning methods and how the overall loss function is formulated and optimized.

\begin{figure*}
  \centering
  \includegraphics[width=0.9\linewidth]{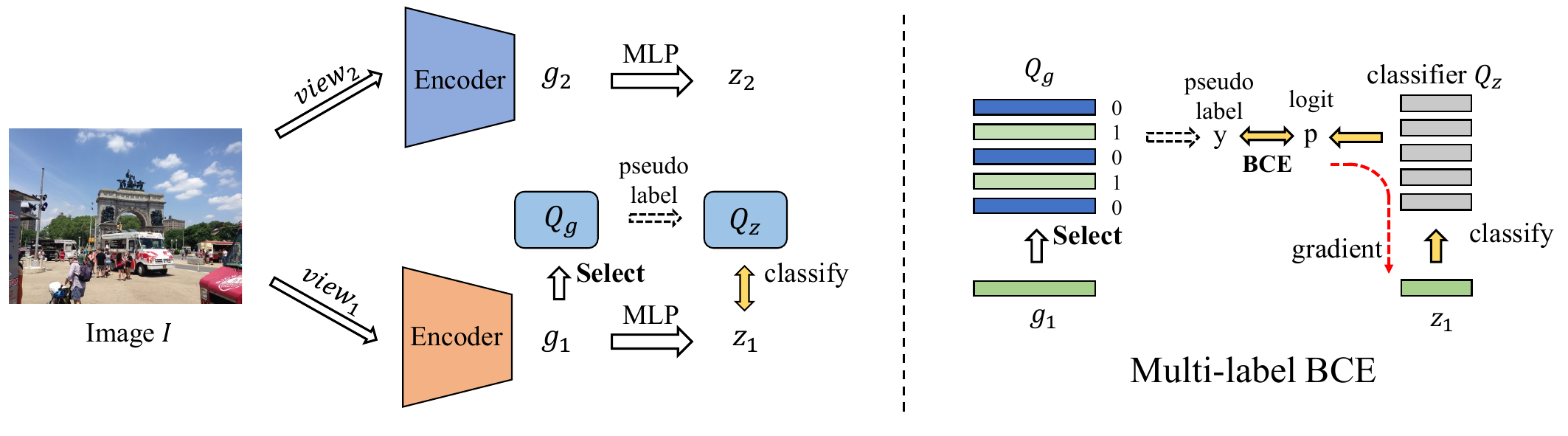}
  \caption{The proposed MLS method. An image $I$ is first augmented to two views and sent to two encoders. The embeddings $g_1,g_2 \in \mathbb{R}^{d_g}$ and $z_1, z_2\in\mathbb{R}^{d_z}$ are features after the backbone and the MLP, respectively. View 2's features ($g_2$ and $z_2$) are enqueued to form two dictionaries $Q_g\in \mathbb{R}^{D\times d_g}$ and $Q_z\in \mathbb{R}^{D\times d_z}$ in the same order. As shown in the right half, the embedding $g_1$ picks its top $k$ nearest embeddings from the queue $Q_g$ to produce binary pseudo labels $y\in \mathbb{R}^D$. Then, the embeddings from $Q_z$ are treated as classifiers that calculate the logit $p\in\mathbb{R}^{D}$ of $z_1$, which is compared against the pseudo-labels $y$ by a BCE loss. Best viewed in color.}
  \label{fig:figure2}
\end{figure*}

\subsection{Preliminaries}

Self-supervised learning methods, whether contrastive based~\cite{MOCO,BYOL} or clustering based~\cite{SwAV}, usually rely on the InfoNCE loss~\cite{InfoNCE} or its variants~\cite{CMC}:
\begin{equation}\label{eq:infoNCE}
\mathcal{L}_{nce} = - \log \frac{\exp{ (q\cdot k_+/\tau)}}
{\exp{(q \cdot k_+/\tau)} + \sum_{k_-} \exp{(q \cdot k_- / \tau)}} \,,
\end{equation}
where $q$ and $k_+$ are positive embeddings (similar to $z_1$ and $z_2$ in Fig.~\ref{fig:figure2}) after the multi-layer perception (MLP) projector of two encoders. The negative features $k_-$ may come from a memory bank~\cite{NPID}, a large dictionary or a queue~\cite{MOCOv2}, or the current mini-batch~\cite{SimCLR}. Negatives may be removed by using additional techniques (\eg, the stop gradient adopted in BYOL~\cite{BYOL} and SimSiam~\cite{Simsiam}).

Similar with the form of a normal softmax function~\cite{CSRA} that is popular in single-label classification, InfoNCE is mutually exclusive among all classes, such that there can only be one positive out of all pseudo concepts. There \emph{are} variants of InfoNCE (such as SupCon~\cite{Supcon}) where more positive pairs are excavated and accumulated, with each item being a sole InfoNCE form. These variants can be summarized as follows:
\begin{equation}
   -\frac{1}{N_{pos}}\sum_{i=1}^{N_{pos}} \log \frac{\exp{ (q\cdot k_+^i/\tau)}} {\exp{(q \cdot k_+^i/\tau)} + \sum_{k_-} \exp{(q \cdot k_- / \tau)}} \,,
\end{equation}
in which the total $N_{pos}$ positives are enumerated over $k_+^i$. It can \emph{indeed} alleviate the dilemma of the traditional form by involving more positives, but all these methods mainly focus on supervised learning~\cite{Supcon} or lack extensive experiments on large dense prediction tasks~\cite{NIPs_revisiting}. 

More importantly, the concept of multi-label learning is \emph{not} introduced, that is, the classes are still mutually exclusive, which hinders the model from learning quality representations from scene image self-supervised learning.

\subsection{Multi-label self-supervised learning}

Now, we will introduce the proposed MLS method in detail. As discussed above, previous SSL methods who targeted scene images all adopt loss functions in the single-label form, while in this paper, we aim to reduce the gap between the optimization objective and the unique property of multi-label images.

We adopt MoCo-v2~\cite{MOCOv2} as our base structure. Specifically, given an input image $I$, it is first cropped and augmented into two different views:
\begin{equation}
   \begin{aligned}
   v_1 = \mathcal{T}(I)\,, \\
   v_2 = \mathcal{T}'(I)\,,
   \end{aligned}
\end{equation}
where $\mathcal{T}$ and $\mathcal{T}'$ are two randomly sampled data augmentations. These two views are then passed through the base encoder $\phi(\cdot)$ and a momentum encoder $\phi^m(\cdot)$ to get the backbone features $g_1$ and $g_2$:
\begin{equation}
       g_1 = \phi(v_1), \, g_2 = \phi^m(v_2)\,,
\end{equation}
in which $g_1,g_2\in \mathbb{R}^{d_g}$ (\eg, $d_g$= 2048 in ResNet-50~\cite{ResNet}). Next, the backbone features are both passed through a multi-layer perceptron (MLP) projector $f(\cdot)$ or $f^m(\cdot)$
\begin{equation}
   z_1 = f(g_1), \, z_2 = f^m(g_2)
\end{equation}
to get the final embeddings $z_1,z_2\in \mathbb{R}^{d_z}$. Similar to MoCo-v2 that retains a large dictionary $Q_z$ (shown in Fig.~\ref{fig:figure2}), here we keep two queues $Q_g\in \mathbb{R}^{D\times d_g}$ and $Q_z \in \mathbb{R}^{D\times d_z}$ with normalized embeddings coming from $g_2$ and $z_2$, respectively. 
\begin{equation}
   \begin{aligned}
       \frac{g_2}{||g_2||} \stackrel{Enqueue}{\Longrightarrow} Q_g \,,\\
       \frac{z_2}{||z_2||} \stackrel{Enqueue}{\Longrightarrow} Q_z \,.
   \end{aligned}
\end{equation}
Note that for simplicity the enqueue and dequeue operations are not shown in Fig.~\ref{fig:figure2}, and that features stored in $Q_g$ and $Q_z$ are all kept in the \emph{same} order. embeddings in both queues are L2 normalized.

Then, the backbone feature $g_1$ of view $v_1$ picks its top $k$ nearest neighbors from $Q_g$ and treats them as containing similar semantics as that in the input image's view $v_1$ (e.g., containing at least one object that appears in the input view $v_1$). These nearest neighbors are given pseudo positive labels and all the rest items in $Q_g$ are pseudo negatives. This step forms multiple binary pseudo-labels $y\in \mathbb{R}^{D}$:
\begin{equation}\label{eq:topk}
   y = IsTopk(g_1\odot Q_g) \,,
\end{equation}
where $\odot$ is the matrix multiplication operation.

In the next step, the normalized embeddings inside $Q_z$ are treated as normalized classifiers. They are multiplied by the feature $z_1$ to produce the logits $p\in \mathbb{R}^D$, which classify $z_1$ into the $D$ pseudo categories:
\begin{equation}
   p = z_1 \odot Q_z \,.
\end{equation}
Finally, the pseudo labels $y$ and the classified logits interact with each other, and are optimized using the binary cross entropy (BCE) loss,
\begin{equation}\label{eq:ml}
   \mathcal{L}_{ml} =  \frac{-1}{D}\sum_{i=1}^D \left[ y_i\log \sigma(\frac{p_i}{\tau}) + (1-y_i) \log (1-\sigma(\frac{p_i}{\tau})\right] \,,
\end{equation}
in which $\tau$ is a temperature hyperparameter (following MoCo-v2's), and $\sigma(\cdot)$ is the sigmoid function~\cite{SRN} that maps the scores to the range $[0,1]$.

\subsection{Optimization}

When we use the multi-label loss $\mathcal{L}_{ml}$ as the \emph{only} loss function, we found that empirically the optimization process often faces an unstable training issue. We believe this unstable optimization should be attributed to the adverse interaction between assigning pseudo-labels and learning representations from a poor starting point. High quality embeddings from the queues $Q_g$ and $Q_z$ cannot be obtained if the pseudo-labels are incorrect. At the same time picking nearest neighbors and classification will surely fail if the enqueued embeddings are misleading (a point which was mentioned in DenseCL~\cite{DenseCL}, too). 

Two strategies easily stabilize the multi-label SSL learning. One is to use InfoNCE alone in the first few epochs to \emph{warmup} the dictionary $Q_g$ and $Q_z$. The other remedy is to combine InfoNCE with our multi-label BCE loss during the whole training process, which is widely used in previous methods~\cite{DenseCL,Piont-Level-recent,SetSim} and easier in implementation. Hence, we adopt this strategy and our overall loss function is formulated as:
\begin{equation}\label{eq:total_loss}
   \mathcal{L} = \mathcal{L}_{nce} + \lambda \mathcal{L}_{ml} \,,
\end{equation}
where $\mathcal{L}_{nce}$ is the InfoNCE loss that is adopted in MoCo-v2 (\cf Eq.~\ref{eq:infoNCE}) and along with many multi-label SSL methods. $\mathcal{L}_{ml}$ is our multi-label BCE loss (\cf Eq.~\ref{eq:ml}). The value of combination weight $\lambda$ is fixed at $\lambda=0.5$. 

The proposed method is abbreviated as MLS (multi-label self-supervised) learning.

\section{Experiments}

Now we validate the effectiveness of MLS through extensive experiments. We first describe the settings of our experiments, including datasets for upstream pretraining and downstream finetuning, training details, and the architecture for object detection, instance/semantic segmentation and image classification.

\subsection{Experimental settings}

\textbf{Datasets.} We adopted the MS-COCO train2017 set~\cite{MS-COCO} for SSL pretraining, which is widely applied in the study of scene image SSL~\cite{ORL,SoCo,DenseCL}. Note that we did \emph{not} use ImageNet. For downstream tasks, we conducted experiments on MS-COCO, VOC0712~\cite{VOC}, CityScapes~\cite{Cityscapes} as well as 7 small classification datasets ( CUB200~\cite{cub200}, Flowers~\cite{flowers}, Cars~\cite{Cars}, Aircraft~\cite{aircrafts}, Indoor67~\cite{indoor}, Pets~\cite{pets} and DTD~\cite{DTD}) following previous SSL methods~\cite{Tobias}. Specifically, MS-COCO contains 118,287 and 5000 images for training and validation, respectively. VOC2007 has 5011 and 4952 images in the trainval and test sets, respectively. CityScapes is a semantic segmentation dataset containing 2975 and 500 images for training and validation, respectively. For more information about the 7 small classification datasets, please refer to~\cite{Tobias} for a more detailed description.

\textbf{Pretraining.} We adopted ResNet-50~\cite{ResNet} as our base encoder (\cf Fig.~\ref{fig:figure2}) and mostly follow the structure of MoCo-v2. We pretrained our models for 400 or 800 epochs on MS-COCO, set learning rate as 0.3, weight decay as 0.0001. For data augmentation, we adopted random resized crop, color distortion, random gray scale, random flip, etc, following~\cite{NIPs_revisiting, SwAV}. The queue size $D$ of $Q_g$ and $Q_z$ was 4096. The temperature $\tau$ was set as 0.2 and was shared between the InfoNCE loss and our multi-label BCE loss. The dimensionalities of the embeddings $g_1$ and $z_1$ were 2048 and 256, respectively. For faster convergence, we adjusted the momentum update coefficient to 0.995. With regard to our multi-label BCE loss, the parameter $k$ was set as 20, and the combination coefficient $\lambda$ was 0.5.

\begin{table*}
  \centering
  \caption{Results of COCO detection and segmentation. All SSL models (except those pretrained on ImageNet) were first pretrained on MS-COCO and then finetuned on COCO with Mask R-CNN R50-FPN~\cite{Mask-RCNN}. Our MLS was pretrained for 400 or 800 epochs for fair comparisons. We reproduced SoCo$^*$ with the settings in~\cite{SoCo}. Note that SetSim~\cite{SetSim} and ReSim~\cite{ReSim} are adaptable to scene image pretraining, but only reported ImageNet pretraining results. The column `Epochs' means the number of pretraining epochs. ORL$^+$~\cite{ORL} needs two pretraining stages (800 epochs each), so the total training time is doubled. }
  \label{tab:COCO-FPN}
  \begin{tabular}{lllllllll}
     \toprule[1pt]
     \multirow{2}{*}{Method} & \multirow{2}{*}{Data} & \multirow{2}{*}{Epochs}   & \multicolumn{3}{c}{Detection} & \multicolumn{3}{c}{Segmentation}  \\

     & & & AP$^{bbox}$ & AP$^{bbox}_{50}$ & AP$^{bbox}_{75}$ & AP$^{seg}$ & AP$^{seg}_{50}$ & AP$^{seg}_{75}$\\
     \midrule[1pt]
      \textcolor{lightgray}{Supervised}~\cite{SoCo} &  \textcolor{lightgray}{ImageNet} &  \textcolor{lightgray}{90}&\textcolor{lightgray}{38.9} &  \textcolor{lightgray}{59.6} &  \textcolor{lightgray}{42.7} &  \textcolor{lightgray}{35.4} &  \textcolor{lightgray}{56.5} &  \textcolor{lightgray}{38.1}   \\
     \midrule
     ReSim-C4~\cite{ReSim} & ImageNet & 200 & 39.3 & 59.7 & 43.1 & 35.7 & 56.7 & 38.1 \\
     LEWEL$_{M}$~\cite{LEWEL} & ImageNet & 200 & 40.0 & 59.8 & 43.7 & 36.1 & 57.0 & 38.7 \\
           SetSim~\cite{SetSim} & ImageNet & 200 & 40.2 & 60.7 & 43.9 & 36.4 & 57.7 & 39.0 \\
     SoCo$^*$~\cite{SoCo} & COCO & 400 & 39.1 & 59.1 & 42.7 & 35.4 & 56.0 & 37.8 \\
     Self-EMD~\cite{self-EMD} &  COCO & 800 & 39.3 & 60.1 & 42.8 & - & - & - \\
     MoCo-v2~\cite{MOCOv2} &  COCO & 800 & 38.5 & 58.1 & 42.1 & 34.8 & 55.3 & 37.3\\
     BYOL~\cite{BYOL} & COCO & 800 & 38.8 & 58.5 & 42.2 & 35.0 & 55.9 & 38.1 \\

           DenseCL~\cite{DenseCL} & COCO & 800 & 39.6 & 59.3 & 43.3 & 35.7 & 56.5 & 38.4 \\
           ORL$^+$~\cite{ORL} & COCO & 1600 & 40.3 & 60.2 & 44.4 & 36.3 & 57.3 & 38.9 \\
           \midrule
           MLS & COCO & 400 & 40.1 & 60.2 & 43.9 & 36.2 & 57.3 & 38.6 \\
           MLS & COCO & 800 & \textbf{40.5} & \textbf{60.7} & \textbf{44.5} & \textbf{36.5} & \textbf{57.7} & \textbf{39.1} \\
     \bottomrule[1pt]
  \end{tabular}
\end{table*}

\textbf{Downstream.} We finetuned the pretrained model on various benchmarks, including MS-COCO object detection and instance segmentation, CityScapes semantic segmentation, VOC0712 detection and other classification benchmarks. Specifically, we finetuned on MS-COCO using the Mask-RCNN R50-FPN/C4~\cite{Mask-RCNN} architecture for 90k iterations. For VOC0712, we finetuned for 24k iterations with Faster-RCNN R50-C4~\cite{faster-rcnn}. For CityScapes semantic segmentation, we adopted PSANet~\cite{PSANet} and PSPNet~\cite{PSPNet} for 40k iterations. For classification, we chose VOC2007 for multi-label recognition and finetuned 120 epochs on 7 small classification datasets. We ran 3 times on VOC detection and 2 runs on CityScapes since these results have relatively large variations. Following previous studies, the evaluation metrics for detection, segmentation and classification are AP, mIoU/mAcc and top-1 accuracy, respectively. All the experiments were conducted with RTX 3090 GPUs in the PyTorch~\cite{pytorch} framework.

\subsection{COCO detection and segmentation results}

We first evaluated downstream performance on MS-COCO object detection and instance segmentation. Following previous work, we used Mask R-CNN R50-FPN for finetuning. The results is in Table~\ref{tab:COCO-FPN}. DenseCL~\cite{DenseCL} and Self-EMD~\cite{self-EMD} are those adopting dense matching, while SoCo~\cite{SoCo} and ORL~\cite{ORL} both resort to the unsupervised object discovery method Selective Search~\cite{SS}. For a fair comparison, we reproduced SoCo's results since it was originally pretrained on COCO+~\cite{SoCo} (which is based on COCO but has more unlabeled data). Note that ReSim-C4~\cite{ReSim} and SetSim~\cite{SetSim} are suitable for scene image pretraining, but only ImageNet results were reported. 

As shown in Table~\ref{tab:COCO-FPN}, with only 400 epochs of pretraining, our MLS is almost better than all counterparts (\eg, DenseCL, SoCo, LEWEL~\cite{LEWEL}) and surpasses supervised results by 1.2\% AP$^{bbox}$ and 0.8\% AP$^{seg}$. With the standard 800 epochs of pretraining, MLS is better than all previous scene image SSL methods. This further verifies our motivation: \emph{formulating scene image self-supervised learning as a multi-label classification is both simpler in concept and effective in performance.}

We also tried the Mask R-CNN R50-C4 structure. Since there are relatively less SSL results based on this structure, we reproduced supervised learning, BYOL~\cite{BYOL}, MoCo-v2~\cite{MOCOv2}, SoCo~\cite{SoCo} and DenseCL~\cite{DenseCL} methods by pretraining and finetuning on MS-COCO. To compare fairly, all these SSL models were pretrained for 400 epochs.

\begin{table}
  \centering
  \small
  \setlength{\tabcolsep}{1.6pt}
  \caption{Results on COCO with the Mask R-CNN R50-C4 detector. `Data' stands for the pretraining dataset. `CC' and `IN' means COCO and ImageNet, respectively. All SSL models were pretrained for 400 epochs on MS-COCO.}
  \label{tab:COCO-C4}
  \begin{tabular}{llllllll}
     \toprule[1pt]
     \multirow{2}{*}{Method} & \multirow{2}{*}{Data}  & \multicolumn{3}{c}{Detection} & \multicolumn{3}{c}{Segmentation}  \\

     & & AP$^{bbox}$ & AP$^{bbox}_{50}$ & AP$^{bbox}_{75}$ & AP$^{seg}$ & AP$^{seg}_{50}$ & AP$^{seg}_{75}$\\

     \midrule
     Random & - &  26.4 & 44.0 & 27.8 & 29.3 & 46.9 & 30.8\\
     Supervised & IN & 38.1 & 58.1& 41.1& 33.2 & 54.8 & 35.0\\
       
     BYOL  & CC & 36.9 & 56.7 & 39.4 & 32.4 & 53.5 & 34.3    \\
     MoCo-v2   & CC & 37.3 & 56.7 & 40.4 & 32.8 & 53.5 & 34.9    \\
           SoCo$^*$ & CC & 36.9 & 56.2 & 39.7 & 32.4 & 53.0 & 34.5\\
     DenseCL & CC & 38.3 & 57.9 & 41.4 & 33.5 & 54.4 & 35.7 \\
           \midrule
           MLS & CC & \textbf{38.6} & \textbf{58.3} & \textbf{41.5} & \textbf{33.8} & \textbf{55.1} & \textbf{36.0} \\
       \bottomrule[1pt]
  \end{tabular}
\end{table}

As can be found in Table~\ref{tab:COCO-C4}, our MLS surpasses all previous self-supervised learning methods like SoCo~\cite{SoCo} and DenseCL~\cite{DenseCL}, despite being much simpler than them. Note that SoCo needs unsupervised box generation and multiple auxiliary loss functions. Our MLS also achieves better results than supervised ImageNet pretraining, with an improvement of 0.5\% AP$^{bbox}$ and 0.6\% AP$^{seg}$ for object detection and instance segmentation, respectively.

\subsection{Transfer learning} 

Next, we test how our MLS method performs on various transfer learning tasks, including CityScapes semantic segmentation, VOC0712 object detection and (multi-label) classification.

\begin{table}
  \setlength{\tabcolsep}{2.5pt}
       \small
  \centering
  \caption{Results of CityScapes semantic segmentation using two different segmentation pipelines (PSANet~\cite{PSANet} and PSPNet~\cite{PSPNet}). `Data' stands for the pretraining dataset. `CC' and `IN' means COCO and ImageNet, respectively.}
  \label{tab:cityscape}
  \begin{tabular}{llllllll}
     \toprule[1pt]
     \multirow{2}{*}{Method} & \multirow{2}{*}{Data}  & \multicolumn{3}{c}{PSANet} & \multicolumn{3}{c}{PSPNet}  \\

     & & mIoU & mAcc & aAcc & mIoU & mAcc & aAcc \\
           \midrule
       Supervised & IN & 77.5 & 86.6 & 95.9 & 77.8 & 86.7 & 95.9 \\
     \midrule
     BYOL& CC & 77.6 & 86.6 & 95.9 & 76.9 & 85.9 & 95.9\\
      MoCo-v2 & CC & 76.8  & 85.8 & 95.8 & 76.8 & 85.4 & 95.8 \\
       DenseCL & CC & 77.6 & 86.6 & 96.0 & 77.6 & 86.3 & 95.9 \\
       SoCo & CC & 76.3 & 85.3 & 95.8 & 76.5 & 85.3 & 95.8 \\

     \midrule
       MLS & CC & \textbf{79.0} & \textbf{87.3} & \textbf{96.2} & \textbf{78.5} & \textbf{86.6} & \textbf{96.1} \\
  \bottomrule[1pt]

  \end{tabular}
\end{table}

\textbf{CityScapes semantic segmentation.} We first transferred pretrained models to CityScapes~\cite{Cityscapes} semantic segmentation. Because relatively few SSL methods worked on this benchmark and the segmentation architecture generally differed, we first reproduced several SSL methods (BYOL, MoCo-v2, DenseCL and SoCo), then finetuned them in the same setting. It is clearly demonstrated in Table~\ref{tab:cityscape} that our multi-label self-supervised learning method is superior than all of them, surpassing state-of-the-art dense matching approach DenseCL~\cite{DenseCL} and object discovery method SoCo~\cite{SoCo}, both by significant margins. Our MLS also improves supervised learning from 77.5\% mIoU to 79.0\% mIoU and from 77.8\% mIoU to 78.5\% mIoU with PSANet~\cite{PSANet} and PSPNet~\cite{PSPNet}, respectively.

\begin{table}
        \centering
  \caption{Results of VOC0712 object detection using the Faster-RCNN R50-C4 object detector. All SSL models were pretrained on MS-COCO for 800 epochs. BYOL and MoCo-v2 results were reproduced by us.}
  \label{tab:VOC-C4}
  \begin{tabular}{lllll}
     \toprule[1pt]

     Method & Data & AP &  AP$_{50}$ & AP$_{75}$ \\
     \midrule
     Random & - & 32.8 & 59.0 & 31.6\\
     Supervised & ImageNet & 53.3 & 81.0 & 58.8 \\
     \midrule
     SwAV~\cite{SwAV}& ImageNet & 45.1 & 77.4 & 46.5 \\
     SimCLR~\cite{SimCLR} & ImageNet & 51.5 & 79.4 & 55.6 \\
           SoCo$^*$~\cite{SoCo} & COCO & 51.7 & 78.6 & 57.2 \\

     Self-EMD~\cite{self-EMD} & COCO & 53.0 &  80.0 & 58.6\\
     BYOL& COCO & 51.7 & 80.2 & 56.4\\
     MoCo-v2 & COCO & 53.7  & 80.0 & 59.5\\
           \midrule
     MLS & COCO & \textbf{55.0} & \textbf{81.6}  & \textbf{61.2}  \\
           
     \bottomrule[1pt]
  \end{tabular}
\end{table}

\textbf{VOC0712 object detection.} Moving on to other dense prediction tasks, we evaluated the effectiveness of MLS on VOC0712 object detection with Faster R-CNN R50-C4. Since there were missing results for BYOL and MoCo-v2 in this experimental setting, we reproduced both of them for a fair comparison. As shown in Table~\ref{tab:VOC-C4}, our MLS surpasses all previous object-centric SSL methods by a large margin. For example, our method is higher than SimCLR by 3.5\% AP and 5.6\% AP$_{75}$ metrics. MLS is also better than \emph{supervised} ImageNet results (an improvement of 1.7\% AP and 2.4\% AP$_{75}$), showing again the great potential of multi-label self-supervised learning.

\textbf{Single- and multi-label classification.} Then, we explore how our proposed method behaves on image classification benchmarks. As pointed out by a previous SSL research work~\cite{DetCo}, various dense matching or unsupervised object discovery based SSL methods usually sacrifice the performance of image classification, in return for high accuracy in dense prediction tasks. In other words, classification is tough for these SSL methods. Therefore, we evaluated MLS on both multi-label (\eg, VOC2007) and single-label classification tasks. As discussed above, we also reproduced the results of MoCo-v2 and DenseCL because MLS has the same architecture with them. It can be found in Table~\ref{tab:classification} that MLS is consistently better than previous SSL methods, surpassing our baseline MoCo-v2 by a noticeable margin---an improvement of 5.3\% mAP for VOC07 and 3.0\% accuracy for Indoor67. We attribute this gain to the joint embedding property of MLS: \emph{since our method is optimized with global features, it can be naturally easier for us to capture image-level information, thus suitable for image classification tasks}. However, all SSL methods are generally worse than supervised ImageNet pretraining, showing that supervised pretraining with more pretraining data (ImageNet vs. COCO) still remains a powerful technique for general image classification.

\begin{table*}
  \centering
  \caption{Results on downstream classification, including VOC2007 multi-label and 7 small single-label classification datasets. We reproduced the baseline MoCo-v2 and dense matching approach DenseCL for a fair comparison. The evaluation metric for VOC2007 is mAP while for others is top-1 accuracy. All models were pretrained for 400 epochs on the MS-COCO dataset.}
  \label{tab:classification}
  \begin{tabular}{llllllllll}
     \toprule[1pt]
       Method & Data & VOC07 &  CUB200 & Flowers & Cars & Aircraft & Indoor & Pets &  DTD \\
       \midrule
       \textcolor{lightgray}{Supervised} & \textcolor{lightgray}{ImageNet} & \textcolor{lightgray}{89.0} & \textcolor{lightgray}{81.3} & \textcolor{lightgray}{96.7} & \textcolor{lightgray}{90.6} &\textcolor{lightgray}{86.7} & \textcolor{lightgray}{58.1} & \textcolor{lightgray}{64.4} & \textcolor{lightgray}{74.7}\\
     \midrule
      MoCo-v2 & COCO & 80.5  & 68.8 & 89.6 & 87.5 & 80.4 & 64.1 & 79.4 & 60.9  \\
       DenseCL & COCO & 83.7 & 69.3 &  88.6 & 88.3 & 79.9 & 65.4 & 80.0 & 61.8\\
       MLS & COCO & \textbf{85.8} & \textbf{71.7} & \textbf{91.2} & \textbf{88.5} & \textbf{81.2} & \textbf{67.1} & \textbf{81.9} & \textbf{63.0}\\

  \bottomrule[1pt]
  \end{tabular}
\end{table*}

\subsection{Ablation studies}

\textbf{Compare with other loss functions.} We first explore if there are other alternatives besides multi-label BCE that can also achieve similar or even better results. An alternative, however, needs to produce \emph{multiple positive paradigms in the loss term}. We consider kNN softmax adopted in~\cite{Supcon,NIPs_revisiting}, and propose a variant of BCE by disregarding the negative term in Eq.~\ref{eq:ml} (calling it `BCE-pos'). As can be seen in Table~\ref{tab:different_bce}, adding kNN softmax and the BCE-pos loss function besides the InfoNCE loss used in~\cite{NIPs_revisiting} both lead to consistent improvements in all metrics, demonstrating the effectiveness of our hypothesis. Still, they are both inferior to our proposed MLS, showing that pure BCE works better than the variant in scene image self-supervised learning.

\begin{figure}
   \centering
   \subfigure[Backbone- and MLP-scores before MLS training]
   {
       \includegraphics[width=0.475\linewidth]{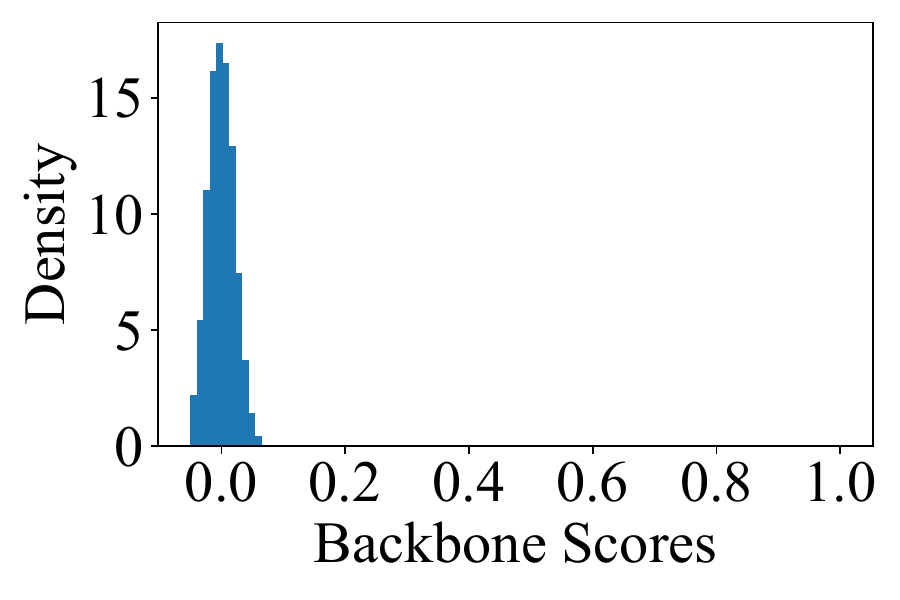}
       \includegraphics[width=0.475\linewidth]{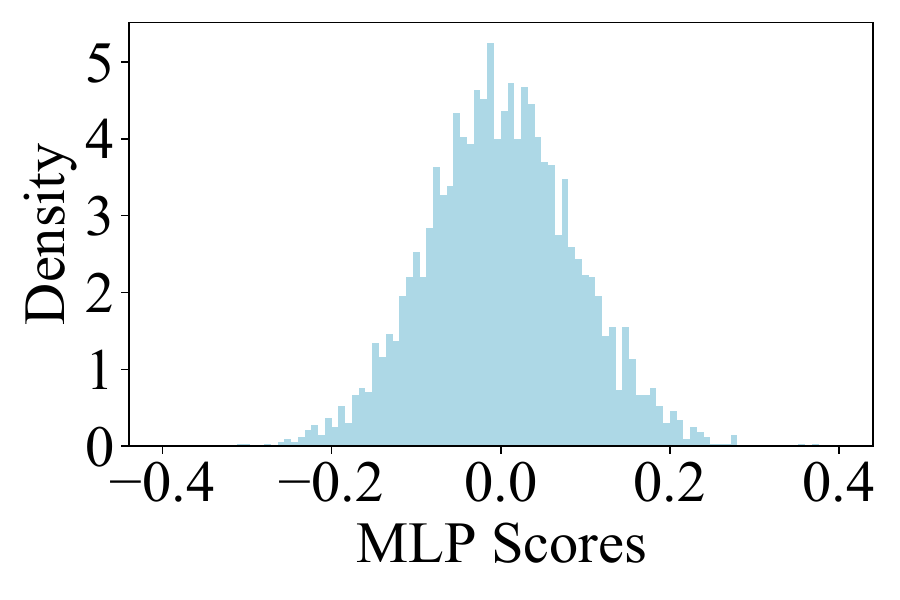}
   }
   \hspace{2pt}
   \subfigure[Backbone- and MLP-scores after MLS training]
   {
       \includegraphics[width=0.475\linewidth]{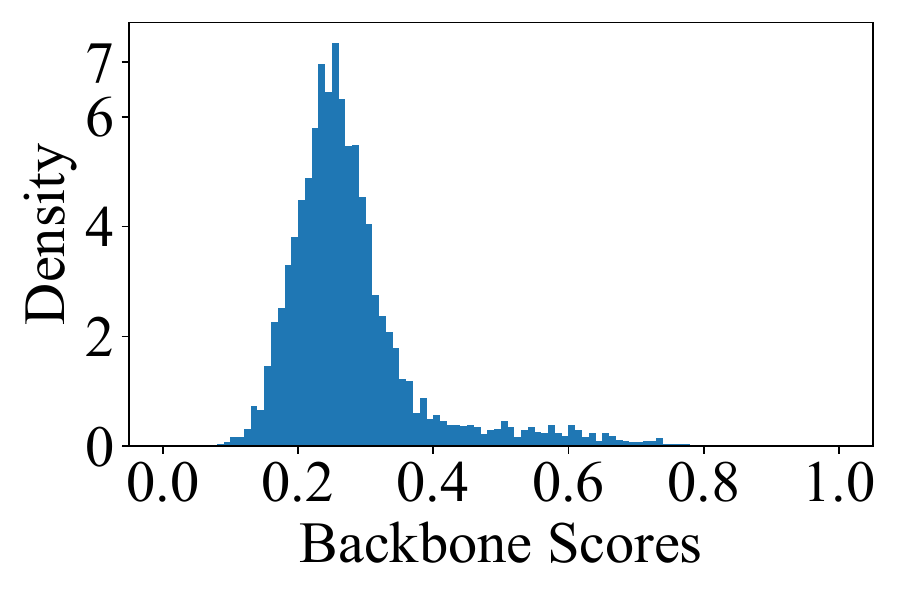}
       \includegraphics[width=0.475\linewidth]{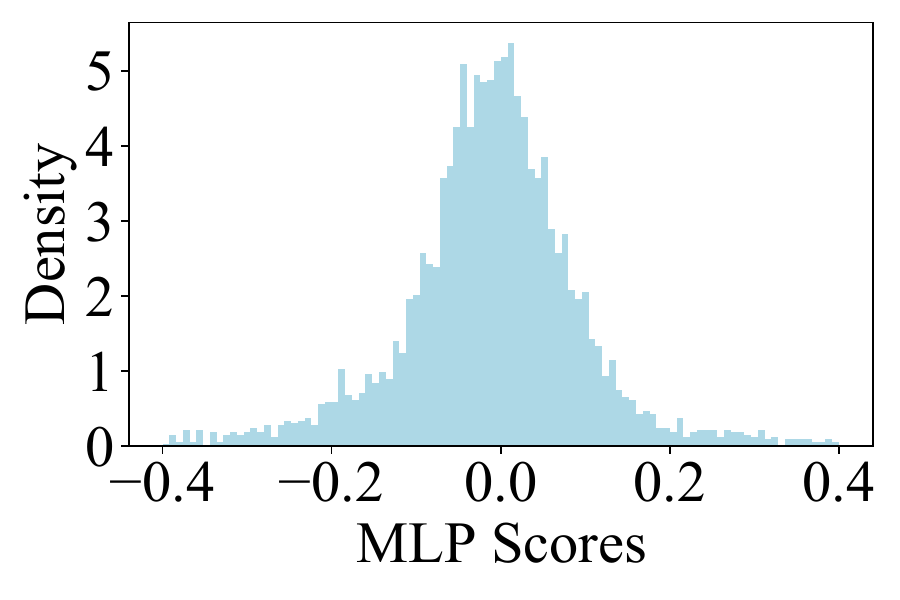}
   }
  \caption{Visualization of the score distribution of one sampled COCO image before and after training. The backbone scores and MLS scores are $g_1 \odot Q_g$ and $z_1\odot Q_z$, respectively. The backbone scores are more indicative compared with MLP scores, thus suitable for pseudo label generation.}
  \label{fig:backbone_score}
\end{figure}

\textbf{Why keeping two dictionaries?} There are two dictionaries $Q_g$ and $Q_z$ in our pipeline with different functionalities: one for pseudo-labeling and the other for classification. Why not keeping only one for simplicity? We explain this problem through both visualization and quantitative experiments. First, we plot the histogram of scores in Fig.~\ref{fig:backbone_score}, with $g_1\odot Q_g$ and $z_1\odot Q_z$ representing the backbone and MLP scores, respectively. The backbone scores have witnessed a dramatic change after training, indicating some information the network potentially obtained by SSL. On the other hand, the MLP scores almost always follow a normal distribution, with only slight changes in the mean and variance statistics. Hence, we choose to use the backbone queue $Q_g$ for pseudo labeling. The MLP dictionary $Q_z$ is used for classification, following the common practice in previous SSL methods~\cite{MOCO,DenseCL}. Our choice is also verified by the experiments in Table~\ref{tab:different_queue}, where using only $Q_g$ or $Q_z$ is clearly sub-optimal, and worse than our MLS where they are jointly adopted.

\begin{table}
  \setlength{\tabcolsep}{2.5pt}
  \small
        \centering
  \caption{MS-COCO detection and segmentation results using different pretraining loss functions. The `kNN' loss was adopted in~\cite{NIPs_revisiting} and the `BCE-pos' loss means only positive terms of binary cross entropy in Eq.~\ref{eq:ml} were considered.}
  \label{tab:different_bce}
  \begin{tabular}{lllllll}
     \toprule[1pt]

     \multirow{2}{*}{Method} & \multicolumn{3}{c}{Detection} & \multicolumn{3}{c}{Segmentation}  \\
        & AP$^{bbox}$ & AP$^{bbox}_{50}$ & AP$^{bbox}_{75}$ & AP$^{seg}$ & AP$^{seg}_{50}$ & AP$^{seg}_{75}$\\
     \midrule
  \textcolor{lightgray}{Baseline} & \textcolor{lightgray}{38.8} & \textcolor{lightgray}{58.3} & \textcolor{lightgray}{42.5} & \textcolor{lightgray}{35.5} & \textcolor{lightgray}{55.5} & \textcolor{lightgray}{37.5}\\
       +kNN & 39.1 &58.7& 42.6& 35.3& 55.8 &37.9 \\
        +BCE-pos & 39.6 & 59.4 & 43.1 & 35.7 & 56.5 & 38.1 \\
  +BCE (\emph{ours}) & \textbf{40.1} & \textbf{60.2} & \textbf{43.9} & \textbf{36.2} & \textbf{57.3} & \textbf{38.6} \\
       \bottomrule[1pt]
  \end{tabular}
\end{table}

\textbf{Effect of $k$.} Now we examine how different $k$ values (the number of pseudo positive labels generated) affect the final results of our MLS method. The downstream results of MS-COCO detection and segmentation can be found in Table~\ref{tab:different_topk}, where a too small and a too large $k$ value both lead to a performance drop. A small $k$ means semantically related images may become false negatives in BCE classification, while a large $k$ will similarly lead to false positives. Our MLS is relatively more robust when $k$ is small. But when $k$ is too large, the accuracy drop is significant. Based on this observation, we choose $k=20$ by default.

\begin{table}
        \centering
  \caption{Effect of using different dictionaries. `$Q_g$ only' means the embeddings in $Q_g$ and $g_1$ are used twice to both generate the pseudo labels $y$ and the logits $p$ (\cf Fig.~\ref{fig:figure2}). `$Q_z$ only' means only using $Q_z$ and $z_1$.}
  \label{tab:different_queue}
  \begin{tabular}{lll}
     \toprule[1pt]
       Method   & AP$^{bbox}$ &AP$^{seg}$ \\
     \midrule
       $Q_g$ only & 39.5 & 35.7 \\
       $Q_z$ only & 39.8 &  36.0 \\
  Both $Q_g$ and $Q_z$ (\emph{ours})  & \textbf{40.1} & \textbf{36.2} \\
       \bottomrule[1pt]
  \end{tabular}
\end{table}

\begin{figure*}
 \centering
 \subfigure
 {
     \includegraphics[width=0.09\linewidth]{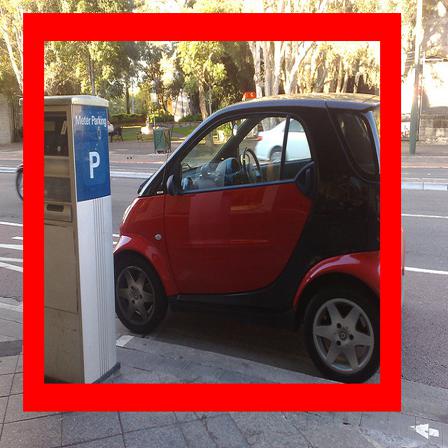}
     \includegraphics[width=0.09\linewidth]{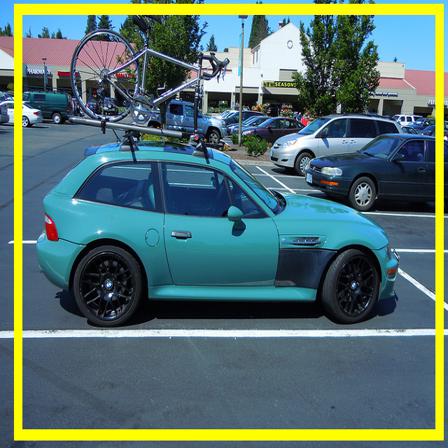}
     \includegraphics[width=0.09\linewidth]{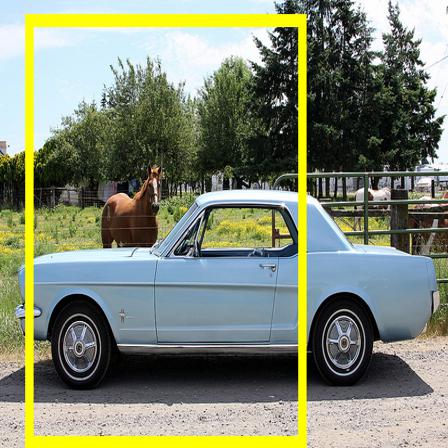}
     \includegraphics[width=0.09\linewidth]{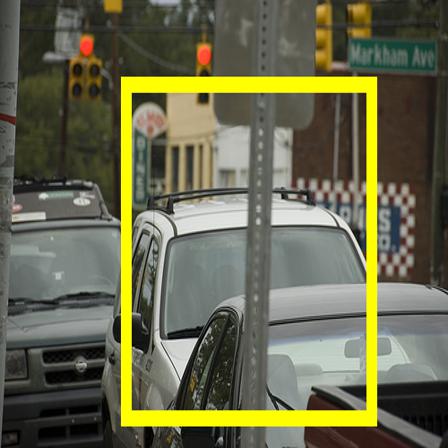}
     \includegraphics[width=0.09\linewidth]{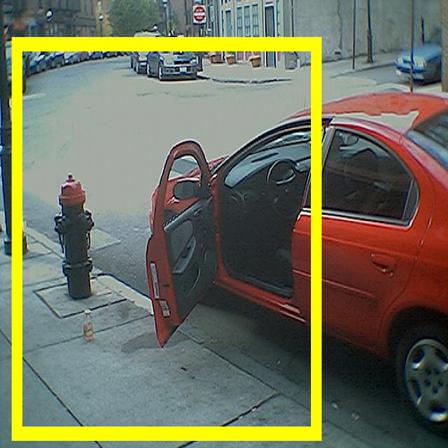}
 } 
 \hspace{10pt}
 \subfigure
 {
     \includegraphics[width=0.09\linewidth]{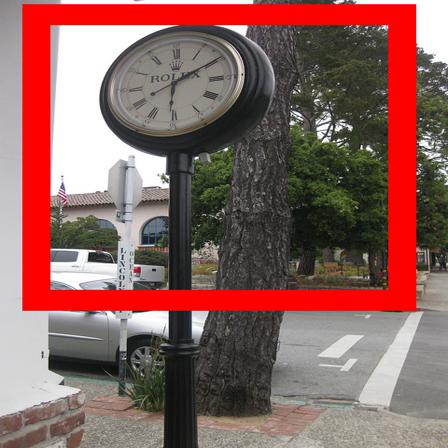}
     \includegraphics[width=0.09\linewidth]{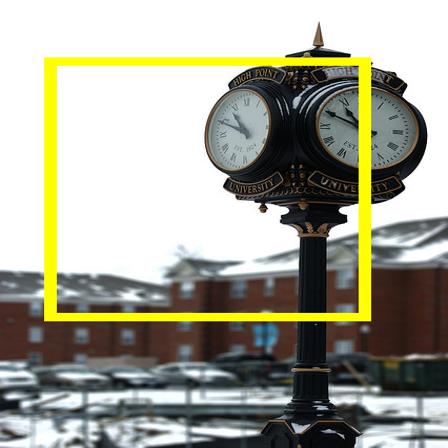}
     \includegraphics[width=0.09\linewidth]{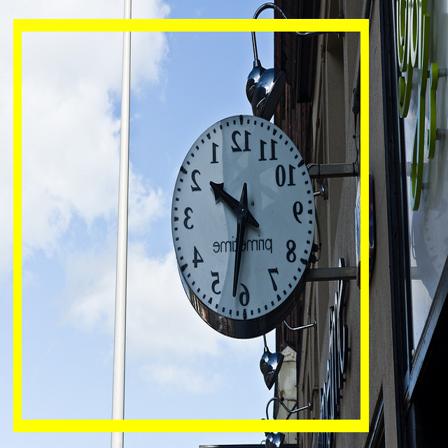}
     \includegraphics[width=0.09\linewidth]{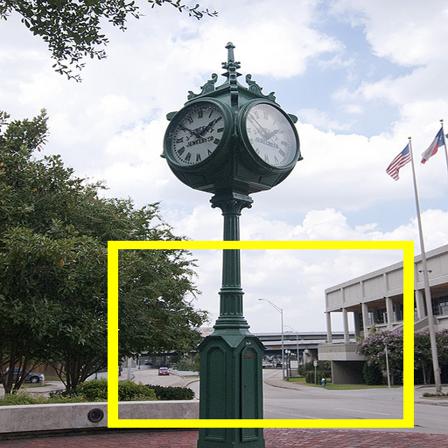}
     \includegraphics[width=0.09\linewidth]{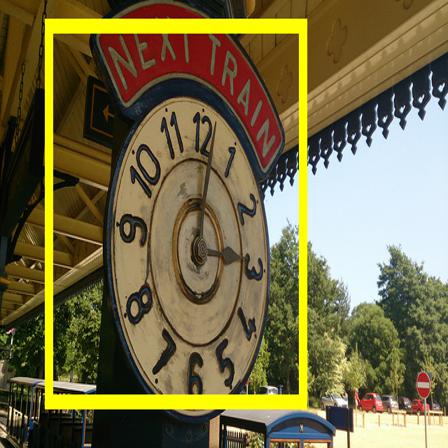}
 }
 \subfigure
 {
     \includegraphics[width=0.09\linewidth]{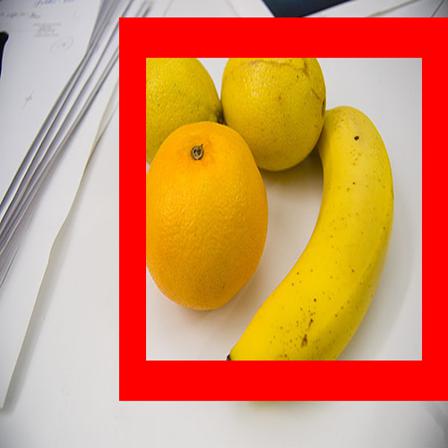}
     \includegraphics[width=0.09\linewidth]{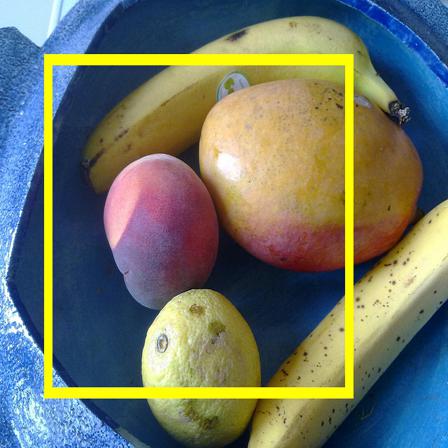}
     \includegraphics[width=0.09\linewidth]{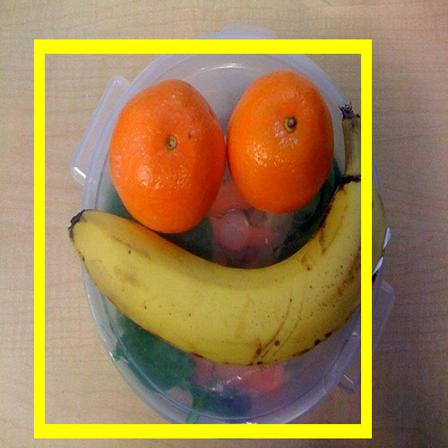}
     \includegraphics[width=0.09\linewidth]{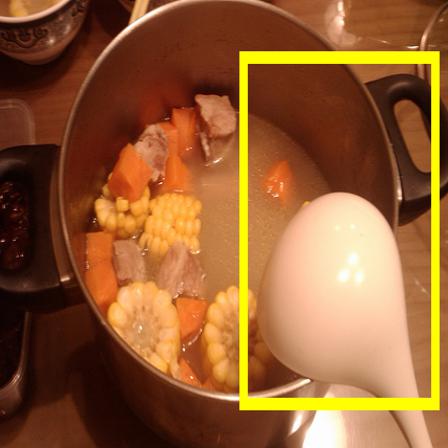}
     \includegraphics[width=0.09\linewidth]{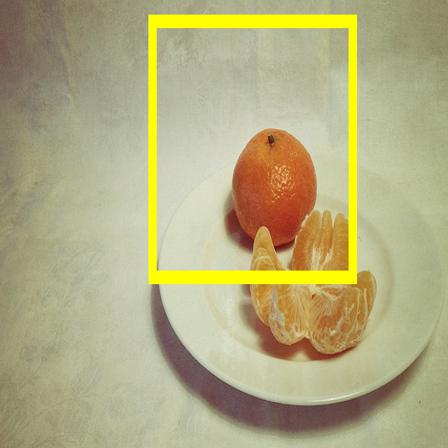}
 } 
 \hspace{10pt}
 \subfigure
 {
     \includegraphics[width=0.09\linewidth]{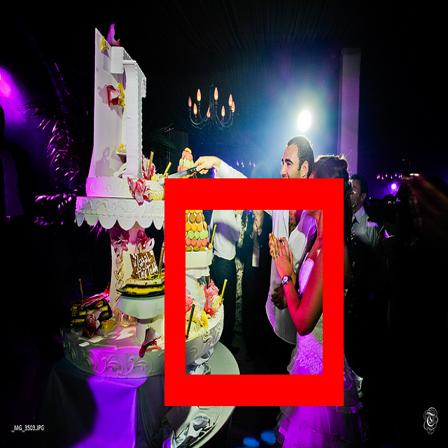}
     \includegraphics[width=0.09\linewidth]{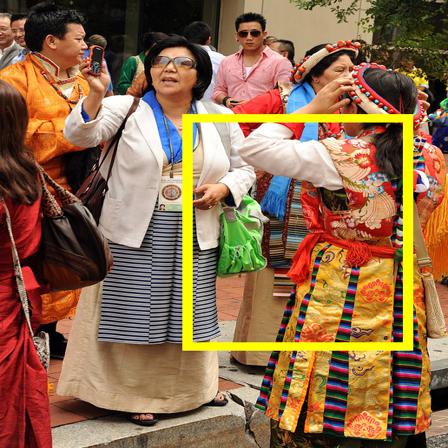}
     \includegraphics[width=0.09\linewidth]{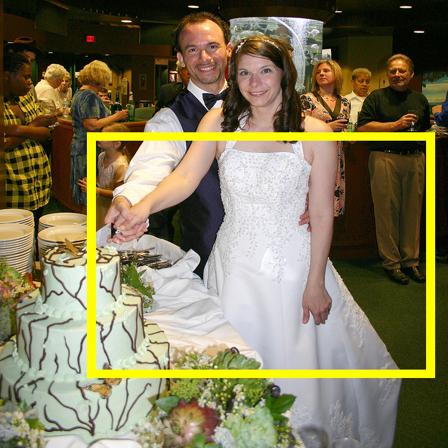}
     \includegraphics[width=0.09\linewidth]{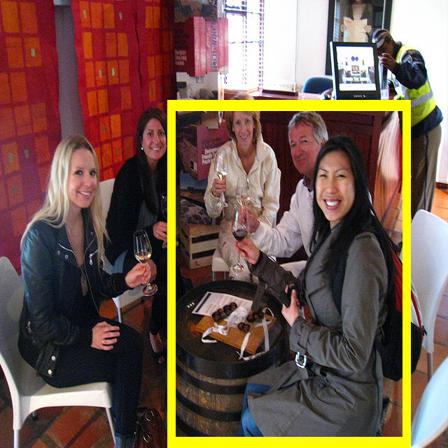}
     \includegraphics[width=0.09\linewidth]{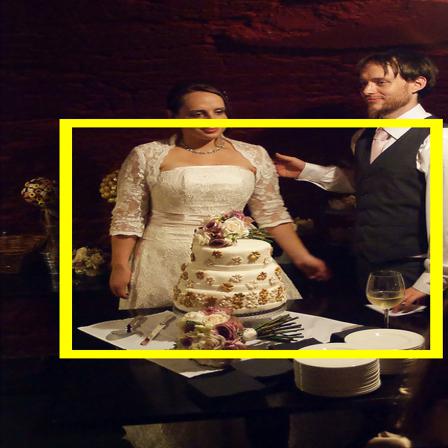}
 }
 \subfigure
 {
     \includegraphics[width=0.09\linewidth]{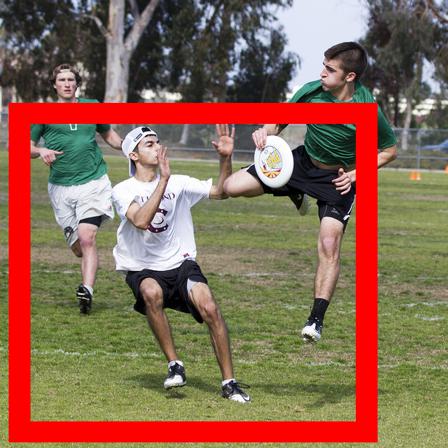}
     \includegraphics[width=0.09\linewidth]{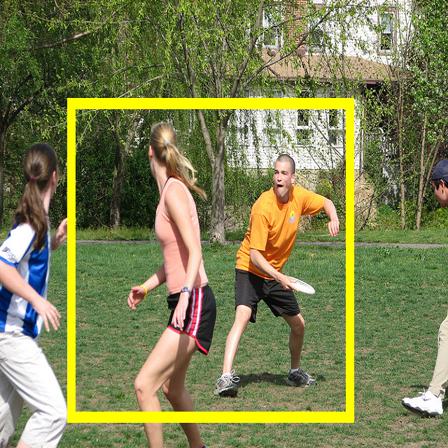}
     \includegraphics[width=0.09\linewidth]{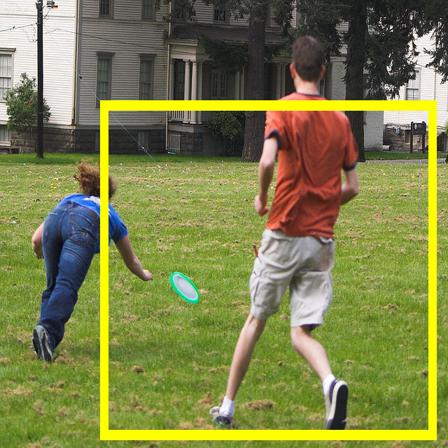}
     \includegraphics[width=0.09\linewidth]{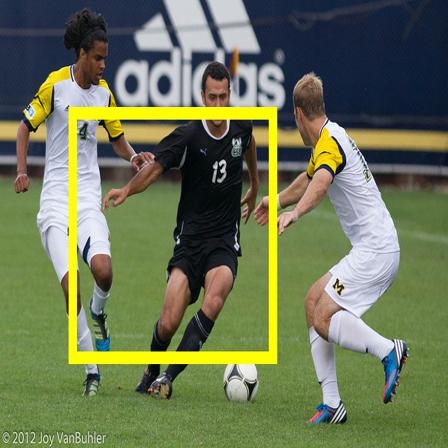}
     \includegraphics[width=0.09\linewidth]{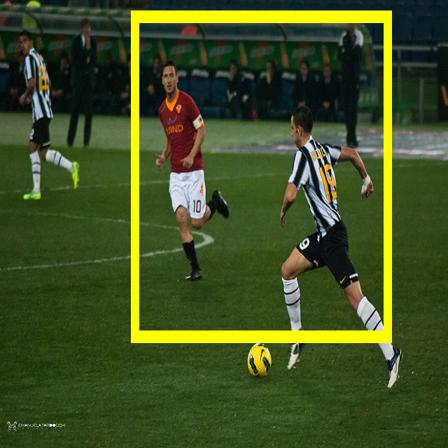}
 }  
 \hspace{10pt}
 \subfigure
 {
     \includegraphics[width=0.09\linewidth]{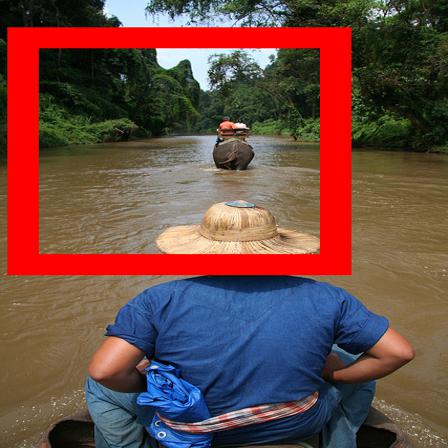}
     \includegraphics[width=0.09\linewidth]{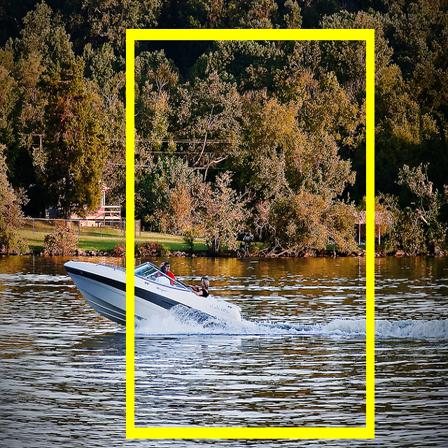}
     \includegraphics[width=0.09\linewidth]{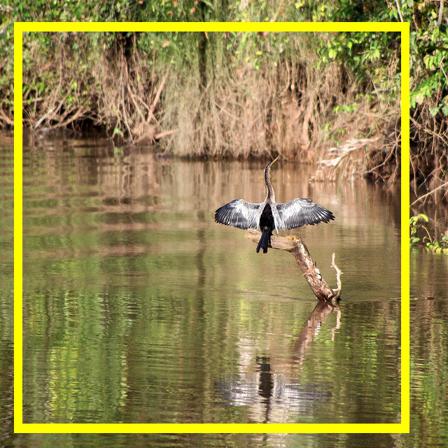}
     \includegraphics[width=0.09\linewidth]{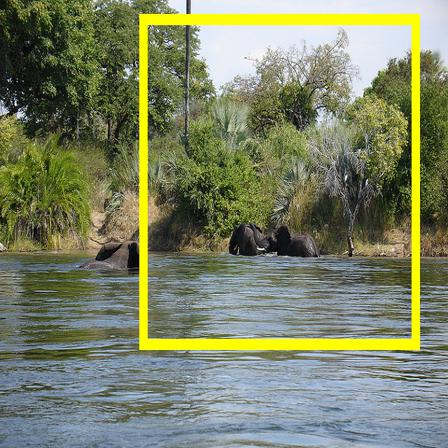}
     \includegraphics[width=0.09\linewidth]{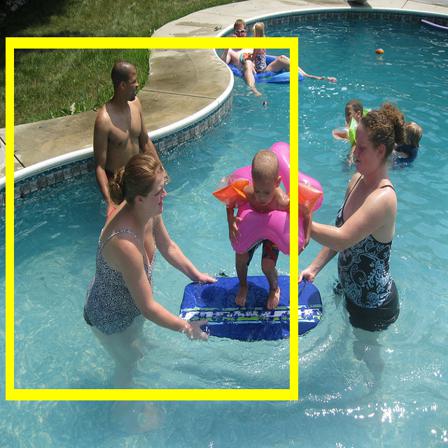}
 }  
 \caption{Visualization of positive pseudo-labels picked up by MLS across images in COCO. For each randomly cropped query image patch, we find the top-4 of its NNs from the dictionary $Q_g$ (shown in yellow rectangles). All these rectangles are generated by the random resized crop data augmentation, a standard component in SSL. We record the cropped boxes' coordinates of the embeddings (the query $g_1$ and those stored in the dictionary $Q_g$ during data augmentation). To be more precise, we first \emph{randomly} sampled 60 \emph{query} images. We then \emph{randomly} sampled 5 query images out of the set of 60 and found their top-4 NNs. In addition, we \emph{manually} select the worst query image (\ie, semantically most different from its top-4 NNs), which is the right half of the last row. This figure is best viewed in color.}
 \label{fig:crosspondence_Visualization}
\end{figure*}

\begin{figure}
   \centering
   \subfigure[Effect of Queue size]
   {
       \includegraphics[width=0.45\linewidth]{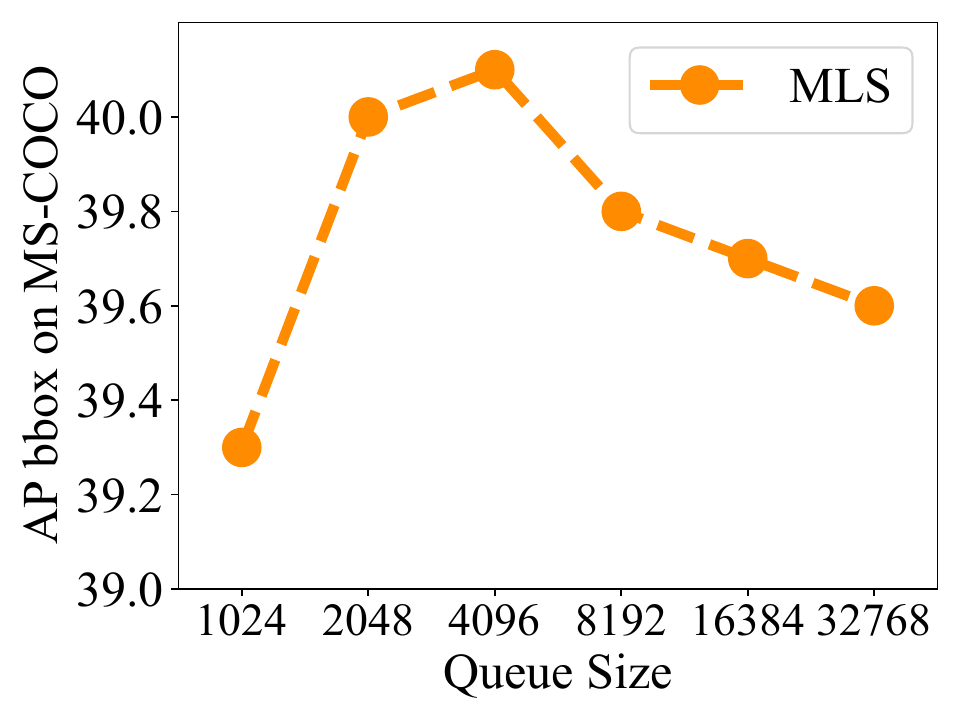}
   }
   \subfigure[Effect of $\lambda$]
   {
       \includegraphics[width=0.45\linewidth]{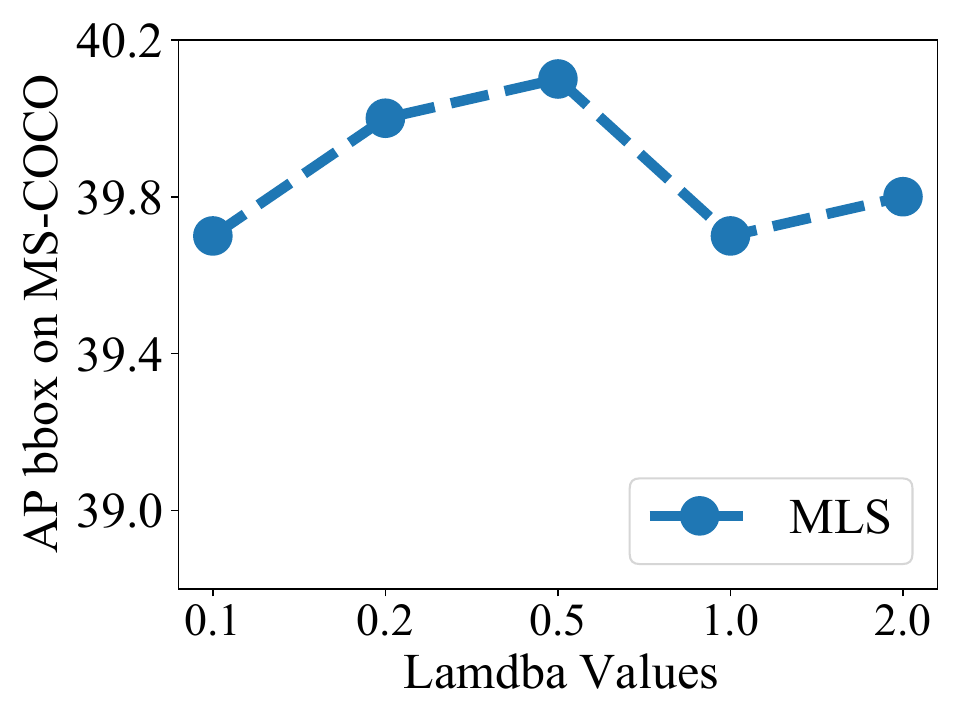}
   }
   \caption{Effect of different queue size and $\lambda$ values (\cf Eq.~\ref{eq:total_loss}) in terms of downstream tasks' results. In this paper, we choose 4096 queue size and $\lambda=0.5$ by default.}
  \label{fig:effect_of_queue_lambda}
\end{figure}

\textbf{Effect of queue size and $\lambda.$} The fourth module we evaluated is the dictionary (queue) size and the combination weight $\lambda$ in Eq.~\ref{eq:total_loss}. We pretrained the models on the MS-COCO dataset using MLS and finetuned them for COCO object detection with Mask R-CNN R50-FPN. As demonstrated in Fig.~\ref{fig:effect_of_queue_lambda} (a), a too small queue size (\eg, 1024) is detrimental for SSL pretraining, since a small queue means limited representations stored, which is not enough for an image to retrieve its positive neighbors. An overly large dictionary also obtains sub-optimal results, because dictionaries that are too large probably contain many out-of-date embeddings (due to the dequeue and enqueue mechanism~\cite{MOCOv2}) and might lead to incorrect pseudo labels. And, we also validate the robustness of our MLS by trying multiple $\lambda$ values. It can be seen in Figure~\ref{fig:effect_of_queue_lambda} (b) that all $\lambda$ leads to consistent improvement over the baseline (\cf Table~\ref{tab:different_bce}), and there is a sweet spot in the choice of $\lambda$. Thus, in this paper we choose $\lambda=0.5$ by default in our experiment.

\begin{table}
         \centering
   \caption{The effect of different $k$ in pseudo label generation as illustrated in Fig.~\ref{fig:figure2} and Eq.~\ref{eq:topk}. All models were pretrained on MS-COCO and finetuned for detection and segmentation.}
   \label{tab:different_topk}
   \begin{tabular}{lllll}
      \toprule[1pt]
        Top $k$  & $k=1$ & $k=5$  & $k=20$ & $k=40$\\
      \midrule
        AP$^{bbox}$ & 39.7 & 40.0  & \textbf{40.1} & 39.2
 \\
        AP$^{seg}$ &35.9 & 36.1 & \textbf{36.2} &  35.4\\
 
        \bottomrule[1pt]
   \end{tabular}
 \end{table}

\textbf{Visualization of positive pseudo-labels.} Last but not least, we validate the effectiveness of our MLS pipeline by visualization: are the top $k$ nearest neighbors (NN) picked up from the dictionary $Q_g$ indeed semantically similar to the input? Because random resized crop is applied in the SSL pipeline (\emph{it is the cropped patch that is sent to the network}), we record the cropped location of each image during data augmentation. Then, for each query $g_1$ in Fig.~\ref{fig:figure2}, we find its top-4 NNs stored in the queue $Q_g$. All these queries and NNs are associated with a certain cropped patch, and thus we visualize them with red and yellow rectangles, respectively. As shown in Fig.~\ref{fig:crosspondence_Visualization}, our MLS backbone effectively captures semantically similar correspondences across the dataset. Specifically, \emph{intra-class variance} (the first row) and \emph{multiple positive partial concepts} (the second and third rows) are clearly demonstrated, showing that our motivation in Sec.~\ref{sec:intro} is valid and that MLS is not only suitable for scene image SSL, but might also be adaptable to image retrieval, as well.

\section{Conclusions and Limitations}

In this paper, we argued that scene image self-supervised learning does not necessarily rely on dense matching or unsupervised object discovery methods, and instead proposed our Multi-Label Self-supervised (MLS) learning approach. The key idea in MLS is that a multi-label image contains multiple concepts or objects, hence we need to have multiple positive pairs corresponding to different objects in the input image. Specifically, we create two dictionaries, one for creating pseudo-labels (positive or negative pairs) and the other for distinguishing between them (\ie, a multi-label classification problem). This is the first time that multi-label SSL is cast as a multi-label classification problem, and MLS has been validated by extensive experiments on various benchmarks (\eg, MS-COCO, CityScapes, VOC0712, etc.) and tasks (object detection, image segmentation, single- and multi-label classification).

As for limitations, it still remains unclear why the BCE loss alone will cause an unstable training issue. This is a long-lasting and open question as numerous scene image SSL methods~\cite{DenseCL,ReSim,SetSim,Piont-Level-recent} also struggle if the InfoNCE loss is removed. Hence, these methods (including our MLS) always use their proposed modules in combination with InfoNCE. In the future, we will explore how and why the collapsing phenomenon exists without the normal contrastive loss, and will apply our MLS to single-label image SSL to further explore the adaptability of our method.

{\small
\bibliographystyle{ieee_fullname}
\bibliography{egbib}
}

\end{document}